\documentclass{article}

\PassOptionsToPackage{numbers, compress}{natbib}

\usepackage[preprint]{neurips_2024}

\usepackage{enumitem}

\usepackage[utf8]{inputenc} 
\usepackage[T1]{fontenc}    
\usepackage{hyperref}       
\usepackage{url}            
\usepackage{booktabs}       
\usepackage{amsfonts}       
\usepackage{nicefrac}       
\usepackage{microtype}      
\usepackage{colortbl}
\usepackage{xcolor}         
\usepackage{natbib}
\usepackage{graphicx}
\usepackage{makecell}
\usepackage{wrapfig,lipsum}
\usepackage{amsmath}
\usepackage{subcaption}
\usepackage[textsize=scriptsize]{todonotes}
\usepackage{multirow}
\usepackage{amsthm}

\usepackage{amsmath,amsfonts,bm}

\def\eqref#1{equation~\ref{#1}}

\def\1{\bm{1}}

\def\valpha{{\bm{\alpha}}}
\def\vbeta{{\bm{\beta}}}

\def\va{{\bm{a}}}
\def\vb{{\bm{b}}}

\def\vh{{\bm{h}}}

\def\vm{{\bm{m}}}

\def\vu{{\bm{u}}}
\def\vv{{\bm{v}}}

\def\vx{{\bm{x}}}

\def\mA{{\bm{A}}}
\def\mB{{\bm{B}}}

\def\mD{{\bm{D}}}

\def\mG{{\bm{G}}}

\def\mK{{\bm{K}}}

\def\mM{{\bm{M}}}

\def\mO{{\bm{O}}}
\def\mP{{\bm{P}}}
\def\mQ{{\bm{Q}}}
\def\mR{{\bm{R}}}

\def\mU{{\bm{U}}}
\def\mV{{\bm{V}}}
\def\mW{{\bm{W}}}

\def\mSigma{{\bm{\Sigma}}}

\DeclareMathAlphabet{\mathsfit}{\encodingdefault}{\sfdefault}{m}{sl}
\SetMathAlphabet{\mathsfit}{bold}{\encodingdefault}{\sfdefault}{bx}{n}

\colorlet{tablerowcolor}{gray!10}
\newcommand{\rowcol}{\rowcolor{tablerowcolor}}

\title{
\begin{minipage}{0.12\textwidth}
  \includegraphics[width=0.85\linewidth]{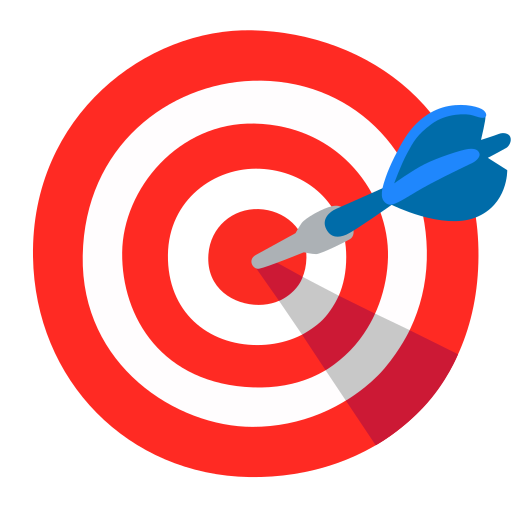}
\end{minipage}
\begin{minipage}{0.85\textwidth}
\textsc{SVFT}: Parameter-Efficient Fine-Tuning \\with Singular Vectors
\end{minipage}
}

\author{%
Vijay Lingam$^{\dagger*}$ \quad Atula Tejaswi$^{\dagger*}$ \quad Aditya Vavre$^{\dagger*}$ \quad Aneesh Shetty$^{\dagger*}$ \\ 
\textbf{Gautham Krishna Gudur}$^{\dagger}$\thanks{indicates equal contribution.} \quad \textbf{Joydeep Ghosh}$^{\dagger}$ \quad \textbf{Alex Dimakis}$^{\dagger}$ \quad \textbf{Eunsol Choi}$^{\dagger}$\\
\textbf{Aleksandar Bojchevski}$^\ddagger$ \quad \textbf{Sujay Sanghavi}$^{\dagger}$ \\
$^\dagger$University of Texas at Austin \quad $^\ddagger$University of Cologne \\
}

\begin{document}

\maketitle

\begin{abstract}

Popular parameter-efficient fine-tuning (PEFT) methods, such as LoRA and its variants, freeze pre-trained model weights \(\mathbf{W}\) and inject learnable matrices \(\mathbf{\Delta W}\). 
These \(\mathbf{\Delta W}\) matrices are structured for efficient parameterization, often using techniques like low-rank approximations or scaling vectors. 
However, these methods typically show a performance gap compared to full fine-tuning. 
Although recent PEFT methods have narrowed this gap, they do so at the cost of additional learnable parameters. 
We propose \textsc{SVFT}, a \textit{simple} approach that fundamentally differs from existing methods: the structure imposed on \(\mathbf{\Delta W}\) depends on the specific weight matrix \(\mathbf{W}\). 
Specifically, \textsc{SVFT} updates \(\mathbf{W}\) as a sparse combination of outer products of its singular vectors, training only the coefficients (scales) of these sparse combinations. 
This approach allows fine-grained control over expressivity through the number of coefficients. 
Extensive experiments on language and vision benchmarks show that \textsc{SVFT}\footnote{code is available at \url{https://github.com/VijayLingam95/SVFT/}} recovers up to \textbf{96\%} of full fine-tuning performance while training only \textbf{0.006 to 0.25}\% of parameters, outperforming existing methods that only recover up to \textbf{85\%} performance using \textbf{0.03 to 0.8\%} of the trainable parameter budget. \looseness=-1

\end{abstract}

\section{Introduction}
\label{sec: introduction}

Large-scale foundation models are often adapted for specific downstream tasks after pre-training. 
Parameter-efficient fine-tuning (PEFT) facilitates this adaptation efficiently by learning a minimal set of new parameters, thus creating an "expert" model. 
For instance, Large Language Models (LLMs) pre-trained on vast training corpora are fine-tuned for specialized tasks such as text summarization~\cite{teaching_machines,pegasus}, sentiment analysis~\cite{t5,liu2019roberta}, and code completion~\cite{codellama} using instruction fine-tuning datasets. 
Although full fine-tuning (Full-FT) is a viable method to achieve this, it requires re-training and storing all model weights, making it impractical for deployment with large foundation models.

To address these challenges, PEFT techniques~\cite{peft} (e.g., LoRA~\cite{lora}) were introduced to significantly reduce the number of learnable parameters compared to Full-FT, though often at the cost of performance. 
DoRA~\cite{dora} bridges this performance gap by adding more learnable parameters and being more expressive than LoRA. Almost all these methods apply a low-rank update additively to the frozen pre-trained weights, potentially limiting their expressivity. 
Furthermore, these adapters are agnostic to the structure and geometry of the weight matrices they modify. 
Finally, more expressive PEFT methods (e.g., LoRA, DoRA, BOFT~\cite{boft}) still accumulate a considerable portion of learnable parameters even in their most efficient configuration (e.g., setting rank=1 in LoRA and DoRA). 
The storage requirements for the learnable adapters can grow very quickly when adapting to a large number of downstream tasks~\cite{vera}.

{Is it possible to narrow the performance gap between \textsc{SVFT} and Full-FT while being highly parameter-efficient?} We propose \textsc{SVFT}: Singular Vectors guided Fine-Tuning — a \textit{simple} approach that involves updating an existing weight matrix by adding to it a sparse weighted combination of {\em its own singular vectors}. The structure of the induced perturbation in \textsc{SVFT} depends on the specific matrix being perturbed, setting it apart from all previous approaches. Our contributions can be summarized as follows:\looseness=-1

\begin{figure*}[t]
\centering
    \includegraphics[width=1.0\linewidth]{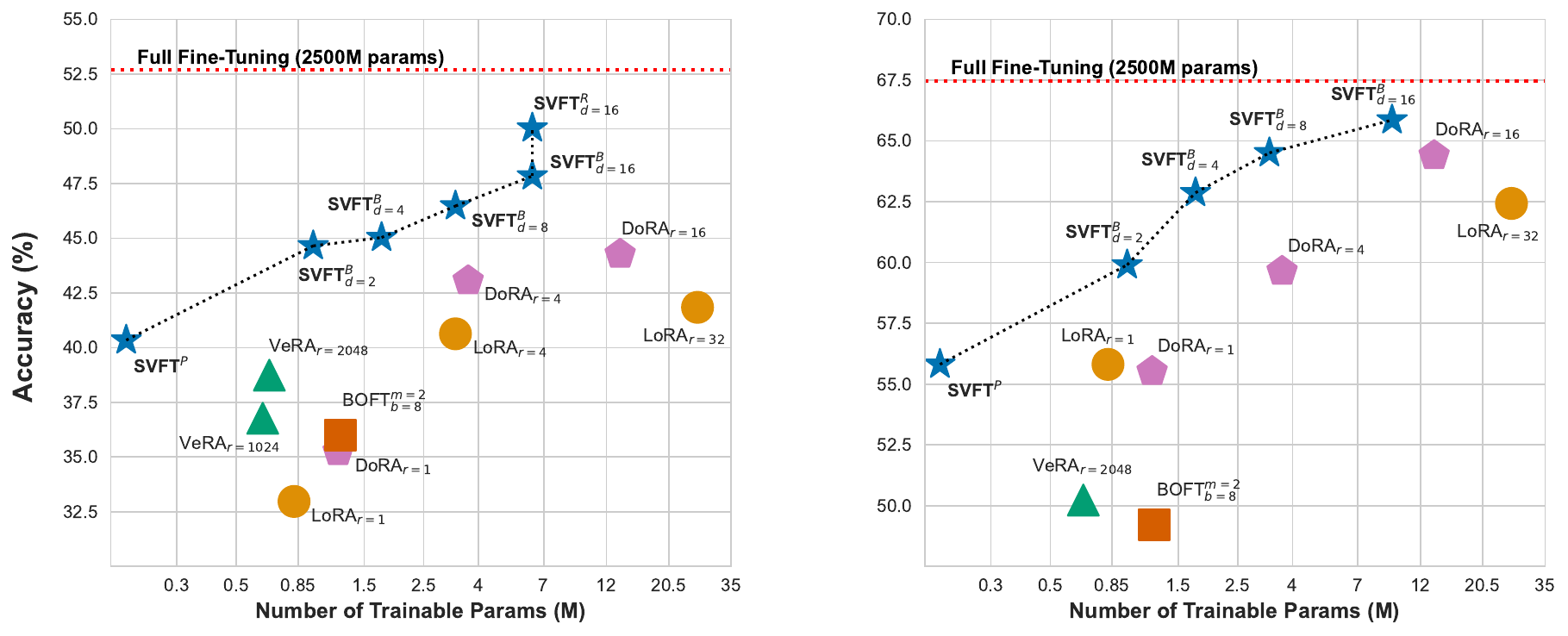}
    \caption{Performance vs total trainable parameters for GSM-8K (left) and Commonsense Reasoning (right) on Gemma-2B. \(\textsc{SVFT}_{d=16}^{B/R}\) outperforms \(\text{DoRA}_{r=8/16}\) with 75\% less trainable parameters.}\looseness=-1
    \label{fig:nlg_plots}
\end{figure*}

\begin{itemize}[leftmargin=5mm]
    \item We introduce \textsc{SVFT}, a new PEFT method. Given a weight matrix \(\mW\), \textsc{SVFT} involves adapting it with a matrix \(\Delta \mW := \sum_{(i,j) \in \Omega} m_{ij} \vu_i \vv_j^T\)  where the \(\{\vu_i\}\) and \(\{\vv_j\}\) are the left and right singular vectors of \(\mW\), \(\Omega\) is an a-priori fixed sparsity pattern, and \(m_{ij}\) for \((i,j)\in \Omega\) are learnable parameters.
    By controlling \(|\Omega|\) we can efficiently explore the accuracy vs parameters trade-off.

    \item \textsc{SVFT} achieves higher downstream accuracy, as a function of the number of trainable parameters, as compared to several popular PEFT methods (see \autoref{fig:nlg_plots}) and over several downstream tasks across both vision and language tasks. Our method recovers up to \textbf{96\%} of full fine-tuning performance while training only \textbf{0.006 to 0.25}\% of parameters, outperforming existing methods that only recover up to \textbf{85\%} performance using \textbf{0.03 to 0.8\%} the trainable parameter budget.
\end{itemize}

We introduce four variants for parameterizing weight updates, namely: \textit{Plain}, \textit{Random}, \textit{Banded}, and \textit{Top-\(k\)} in \textsc{SVFT} (which differ in their choices of the fixed sparsity pattern \(\Omega\)) and validate these design choices empirically. Additionally, we theoretically show that for any fixed parameters budget, \textsc{SVFT} can induce a higher rank perturbation compared to previous PEFT techniques.

\section{Related Work}

Recent advancements in large language models (LLMs) have emphasized the development of PEFT techniques to enhance the adaptability and efficiency of large pre-trained language models.

\textbf{LoRA.} A notable contribution in this field is Low-Rank Adaptation (LoRA)~\cite{lora}, which freezes the weights of pre-trained models and integrates trainable low-rank matrices into each transformer layer.

For a pre-trained weight matrix \(\mW_0\in \mathbb{R}^{d\times n}\), LoRA constraints the weight update \(\Delta \mW\) to a low-rank decomposition: \(\vh = \mW_0\vx + \Delta \mW \vx = \mW_0\vx + \underline{\mB\mA}\vx\), where \(\mB\in \mathbb{R}^{d\times r}\), \(\mA\in \mathbb{R}^{r\times n}\) and rank \(r\ll\min(d,n)\). We underline the (trainable) parameters that are updated via gradient descent.

\textbf{LoRA variants.} We highlight some recent approaches that further improve the vanilla LoRA architecture. Vector-based Random Matrix Adaptation (VeRA)~\cite{vera} minimizes the number of trainable parameters by utilizing a pair of low-rank random matrices shared between layers and learning compact scaling vectors while maintaining performance comparable to LoRA. Formally, VeRA can be expressed as: \(\vh = \mW_0\vx + \Delta \mW \vx = \mW_0 \vx + \underline{\mathbf{\Lambda}_b}\mB\underline{\mathbf{\Lambda}_d}\mA \vx\), where \(\mA\) and \(\mB\) are initialized randomly, frozen, and shared across layers, while \(\mathbf{\Lambda}_b\) and \(\mathbf{\Lambda}_d\) are trainable diagonal matrices.
\looseness=-1

An alternative approach, Weight-Decomposed Low-Rank Adaptation (DoRA)~\cite{dora}, decomposes pre-trained weight matrices into magnitude and direction components, and applies low-rank updates for directional updates, reducing trainable parameters and enhancing learning capacity and training stability. DoRA can be expressed as: \(\vh = \underline{\vm}\frac{\mW_0+\Delta \mW}{\|\mW_0+\Delta \mW\|_c}\vx = \underline{\vm}\frac{\mW_0+\underline{\mB\mA}}{\|\mW_0+\underline{\mB\mA}\|_c}\vx\), where \(\|\cdot \|_c\) denotes the vector-wise norm of a matrix across each column. Similar to LoRA, \(\mW_0\) remains frozen, whereas the magnitude vector \(\vm\) (initialized to \(\|\mW_0\|_c\)) and low-rank matrices \(\mA, \mB\) contain trainable parameters.\looseness=-1

AdaLoRA~\cite{adalora} adaptively distributes the parameter budget across weight matrices based on their importance scores and modulates the rank of incremental matrices to manage this allocation effectively. PiSSA (Principal Singular Values and Singular Vectors Adaptation)~\cite{pissa} is another variant of LoRA, where matrices \(\mA, \mB\) are initialized with principal components of SVD and the remaining components are used to initialize \(\mW_0\). FLoRA~\cite{flora} enhances LoRA by enabling each example in a mini-batch to utilize distinct low-rank weights, preserving expressive power and facilitating efficient batching, thereby extending the domain adaptation benefits of LoRA without batching limitations. \looseness=-1

\textbf{Other PEFT variants.} Orthogonal Fine-tuning (OFT)~\cite{oft} modifies pre-trained weight matrices through orthogonal reparameterization to preserve essential information. However, it still requires a considerable number of trainable parameters due to the high dimensionality of these matrices. Butterfly Orthogonal Fine-tuning (BOFT)~\cite{boft} extends OFT's methodology by incorporating Butterfly factorization thereby positioning OFT as a special case of BOFT. Unlike the additive low-rank weight updates utilized in LoRA, BOFT applies multiplicative orthogonal weight updates, marking a significant divergence in the approach but claims to improve parameter efficiency and fine-tuning flexibility. BOFT can be formally expressed as: \(\vh = (\underline{\mR(m,b)}\cdot \mW_0)\vx\), where the orthogonal matrix \(\mR(m,b)\in\mathbb{R}^{d\times d}\) is composed of a product of multiple orthogonal butterfly components. When \(m=1\), BOFT reduces to block-diagonal OFT with block size \(b\). When \(m=1\) and \(b=d\), BOFT reduces to the original OFT with an unconstrained full orthogonal matrix.

\begin{figure*}[t]
\centering
    \includegraphics[width=1.0\linewidth]{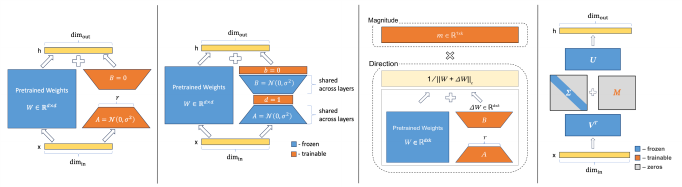}
    \caption{Schematic comparison of LoRA, VeRA, DoRA, and \textsc{SVFT} (left to right).}
    \label{fig:svft_relatedworks}
\end{figure*}

\section{Method}
\label{sec:method}

In this section, we introduce Singular Vectors guided Fine-Tuning (\textsc{SVFT}). The main innovation in \textsc{SVFT} lies in applying structure/geometry-aware weight updates.  

\subsection{\textsc{SVFT} Formulation}
We now formally describe our method, \textsc{SVFT} for parameter-efficient fine-tuning of a pre-trained model. Let \(\mW_{0} \in \mathbb{R}^{d_1 \times d_2}\) denote a weight matrix in the pre-trained model. For instance, in a transformer block, this could be the key matrix, the query matrix, a matrix in the MLP, etc. We add a structured, learned \(\Delta \mW\) to this matrix as follows.

As a first step, we compute the Singular Value Decomposition (SVD) of the given matrix: \(\mW_{0}= \mU \mSigma \mV^{T}\). That is, \(\mU\) is the \(d_1 \times d_1\) matrix of left singular vectors (i.e., its columns are orthonormal), \(\mV^T\) is the \(d_2 \times d_2\) matrix of right singular vectors (i.e., its rows are orthonormal), and \(\mSigma\) is a \(d_1\times d_2\) diagonal matrix. Then, we parameterize our weight update as  
\(\Delta \mW ~ = ~ \mU {\underline{\mM}} \mV^T\), where \(\mU, \mV\) are fixed and frozen, while \(\underline{\mM}\) is a \(d_1 \times d_2\) {\bf sparse trainable matrix with pre-determined and fixed sparsity pattern}\footnote{Learnable parameters are underlined.}. That is, we first pre-determine a small fixed set of elements in \(\mM\) that will be allowed to be non-zero and train only those elements. The forward pass for \textsc{SVFT} can be written as,
\begin{equation}
    \vh = \mW_0 x + \Delta \mW x =   \mU (\mSigma + \underline{\mM}) \mV^T \vx
\label{eq:main_eq}
\end{equation}

\begin{figure*}[t]
\centering
    \includegraphics[width=1.0\linewidth]{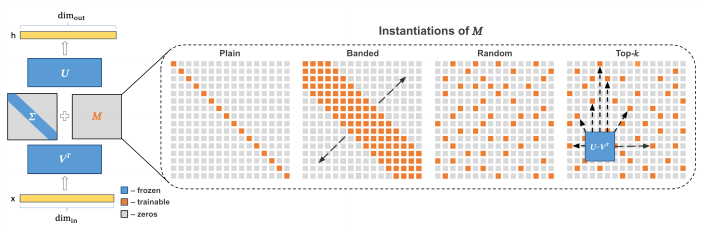}
    \caption{An Overview of \textsc{SVFT}. The original weights \(\mW\) are decomposed into \(\mU, \mathbf{\Sigma}, \mV\). Here, \(\mM\) contains all the trainable parameters, which can be configured into patterns such as Plain, Random, Banded, and Top-\(k\), represented by patterns of trainable (orange) and zero (gray) elements.}
    \label{fig:svft_diagram}
\end{figure*}

We explore four choices for \(\Omega\), the a-priori fixed sparsity pattern of \(\underline{\mM}\). \\
\textbf{Plain} \(\left(\textsc{SVFT}^{P}\right)\). In this variant, we constrain \(\underline{\mM}\) to be a diagonal matrix, which can be interpreted as adapting singular values and reweighting the frozen singular vectors. Since only the diagonal elements are learned, this is the most parameter-efficient \textsc{SVFT} variant. \\
\textbf{Banded} \(\left(\textsc{SVFT}_{d}^{B}\right)\). In this approach, we populate \(\underline{\mM}\) using a banded matrix, progressively making off-diagonals learnable. Specifically, for constants \(z_1\) and \(z_2\), \(\underline{\mM_{ij}}=0\) if \(j < i - z_1\) or \(j > i + z_2\), where \(z_1, z_2 \geq 0\). In our experiments, we set \(z_1 = z_2 = d\) to induce off-diagonal elements that capture additional interactions beyond those represented by singular values. This banded perturbation induces local interactions, allowing specific singular values to interact with their immediate neighbors, ensuring smoother transitions. This method, although deviating from the canonical form of SVD, provides a mechanism to capture localized interactions.\\
\textbf{Random} \(\left(\textsc{SVFT}_{d}^{R}\right)\). A straightforward heuristic for populating \(\underline{\mM}\) involves randomly selecting \(k\) elements to be learnable.\\
\textbf{Top-\(k\)} \(\left(\textsc{SVFT}_{d}^{T}\right)\). The final design choice we explore involves computing the alignment between the left and right singular vectors as \(\vu_i^{T} \vv_j\). We then select the top-\(k\) elements and make them learnable. However, note that this only works when left and right singular vectors have the same size. 
A possible interpretation of this is we make only the top-\(k\) strong interactions between singular vector directions learnable.\looseness=-1

We illustrate these \textsc{SVFT} design choices in \autoref{fig:svft_diagram}. 
Our empirical results demonstrate that these simple design choices significantly enhance performance compared to state-of-the-art PEFT methods. Note that \(\textsc{SVFT}^{P}\) has a fixed number of learnable parameters, while the remaining variants are configurable. We hypothesize that further innovation is likely achievable through optimizing the sparsity pattern of \(\underline{\mM}\), including efficient learned-sparsity methods. In this paper, we explore these four choices to validate the overall idea: determining a perturbation using the singular vectors of the matrix that is being perturbed.

\subsection{Properties of \textsc{SVFT}}
\label{subsection:insights}

We highlight some properties of \textsc{SVFT} in the following lemma and provide insights into how its specific algebraic structure compares and contrasts with baseline PEFT methods.

{\bf Lemma:} Let \(\mW_0\) be a matrix of size \(d_1 \times d_2\) with SVD given by \(\mU \mSigma \mV^T\). 
Consider an updated final matrix \(\mW_0 + \mU\mM \mV^T\), where \(\mM\) is a matrix of the same size as \(\mSigma\), which may or may not be diagonal. 
Then, the following holds: 
\begin{enumerate}
\item[{\em (a)}] {\em Structure:} If \(\mM\) is also diagonal (i.e. the plain \textsc{SVFT}), then the final matrix \(\mW_0 + \mU \mM \mV^T\) has \(\mU\) as its left singular vectors and \(\mathrm{sign}(\mSigma + \mM )\mV^T\) as its right singular vectors. That is, its singular vectors are unchanged, except for possible sign flips. 
Conversely, if \(\mM\) is {\em not} diagonal (i.e., variants of \textsc{SVFT} other than plain), then \(\mU\) and \(\mV\) may no longer be the singular directions of the final matrix \(\mW_0 + \mU \mM \mV^T\). 

\item[{\em (b)}] {\em Expressivity:} Given {\em any} target matrix \(\mP\) of size \(d_1 \times d_2\), there exists an \(\mM\) such that \(\mP = \mW_0 + \mU \mM \mV^T\). 
That is, if \(\mM\) is fully trainable, any target matrix can be realized using this method.

\item[{\em (c)}]{\em Rank:} If $\mM$ has $k$ non-zero elements, then the rank of the update \(\mU \mM \mV^T\) is at most \(\min\{k,\min\{d_1,d_2\}\}\).
For the same number of trainable parameters, \textsc{SVFT} can produce a much higher rank perturbation than LoRA (eventually becoming full rank), but in a constrained structured subspace. \looseness=-1

\end{enumerate}

We provide our proofs in Appendix \ref{app:proofs}. Building on this lemma, we now compare the form of the \textsc{SVFT} update with LoRA and VeRA.

\textsc{SVFT}'s \(\Delta \mW\) can be written as a sum of rank-one matrices:

\begin{equation}
\label{eq:2}
    \Delta \mW ~ = ~ \sum_{(i,j)\in \Omega } {\underline{ m_{ij}}} \vu_i \vv_j^T
\end{equation}

where \(\vu_i\) is the \(i^{th}\) left singular vector, \(\vv_j\) is the \(j^{th}\) right singular vector, and \(\Omega\) is the set of non-zero elements in \(\mM\). 

Thus, our method involves adding a weighted combination of specific rank-one perturbations of the form \(\vu_i \vv_j^T\).

LoRA and VeRA updates can also be expressed as sums of rank-one matrices.
\begin{equation}
\label{eq:3}
    \Delta \mW_{\text{LoRA}} ~ = ~ \sum_{i=1}^r  {\underline{\va_i} ~ \underline{\vb_i}^T} 
    \quad \text{and} \quad
    \Delta \mW_{\text{VeRA}} ~ = ~ \sum_{i=1}^r  {\underline{\alpha_i}} (\hat{\va}_i \odot  {\underline{\vbeta}} ) \hat{\vb}_i^T
\end{equation}

where \(\underline{\va_i}\) and \(\underline{\vb_j}\) are the trainable columns of \(\mA\) and \(\mB\) matrices in LoRA. In VeRA, \(\hat{\va}_i\) and \(\hat{\vb}_i\) are random and fixed vectors, while \(\underline{\valpha}\) and \(\underline{\vbeta}\) represent the diagonal elements of \(\mathbf{\Lambda}_d\) and \(\mathbf{\Lambda}_b\) respectively.

Note that LoRA requires \(d_1 + d_2\) trainable parameters per rank-one matrix, while \textsc{SVFT} and VeRA require only one. 
Although LoRA can potentially capture directions different from those achievable by the fixed \(\{\vu_i, \vv_j^T\}\) pairs, each of these directions incurs a significantly higher parameter cost.

VeRA captures new directions at a parameter cost similar to \textsc{SVFT}; however, there is a key distinction: in VeRA, each vector \(\hat{\va}_i\) or \(\hat{\vb}_i\) appears in only one of the rank-one matrices. In contrast, in \textsc{SVFT}, the same vector \(\vu_i\) can appear in multiple terms in the summation, depending on the sparsity pattern of \(\mM\). This results in an important difference: unlike \textsc{SVFT}, VeRA is {\em not universally expressive} -- it cannot represent any target matrix \(\mP\). Moreover, \(\hat{\va}_i, \hat{\vb}_i\) are random, while \(\vu_i, \vv_j\) depend on \(\mW_0\).

\textbf{Note.} \textsc{SVFT} requires storing both left and right singular vectors due to its computation of the SVD on pre-trained weights. While this increases memory usage compared to LoRA (which is roughly double), it remains lower than BOFT. We partially address this through system-level optimizations like mixed-precision weights (e.g., bfloat16). Further exploration of memory-reduction techniques, such as quantization, is planned as future work. Importantly, inference time and memory consumption remain the same across all methods, including \textsc{SVFT}, as the weights can be fused.

\section{Experiments}
\label{sec:experiments}

\subsection{Base Models}
\label{sec:base_models}

We adapt widely-used language models, encoder-only model (DeBERTaV3\textsubscript{base}~\cite{debertav3}) and two decoder-only models (Gemma-2B/7B~\cite{gemma}, LLaMA-3-8B~\cite{llama3}). We also experiment with vision transformer models (ViT-B/16 and ViT-L/16)~\cite{vit}) pre-trained on ImageNet-21k~\cite{imagenet}, following prior work~\cite{vera}. The complete details of our experimental setup and hyperparameter configurations are provided in \autoref{app:additional_exp}. \\
\textbf{Baselines.} We compare with \textbf{Full Fine-Tuning (FT)} updating all learnable parameters in all layers, along with \textbf{LoRA}~\cite{lora},  \textbf{DoRA}~\cite{dora}, \textbf{BOFT}~\cite{boft} and \textbf{VeRA}~\cite{vera}.\footnote{BOFT is approximately three times slower than LoRA. The shared matrices in VERA can become a limiting factor for models with non-uniform internal dimensions, such as LLaMA-3.}

\subsection{Datasets}
\label{sec:datasets}

\textbf{Language.} For natural language generation (NLG) tasks, we evaluate on GSM-8K~\cite{gsm8k} and MATH~\cite{hendrycks2021measuring} by fine-tuning on MetaMathQA-40K~\cite{yu2023metamath}, following~\cite{boft}. We also evaluate on 8 commonsense reasoning benchmarks (BoolQ~\cite{boolq}, PIQA~\cite{piqa}, SIQA~\cite{socialiqa}, HellaSwag~\cite{hellaswag}, Winogrande~\cite{winogrande}, ARC-easy/challenge~\cite{arc}, and OpenBookQA~\cite{openbookqa}). We follow the setting outlined in prior work~\cite{dora,hu2023llmadapters}, where the training sets of all benchmarks are amalgamated for fine-tuning. We fine-tune on 15K examples from this training set. For natural language understanding (NLU), we evaluate on the  General Language Understanding Evaluation (GLUE) benchmark consisting of classification and regression tasks, in line with~\cite{vera, lora}. \\
\textbf{Vision.} Our experiments on vision tasks consist of 4 benchmarks:  CIFAR-100~\cite{krizhevsky}, Food101~\cite{food101}, RESISC45~\cite{resisc45}, and Flowers102~\cite{flowers102}. We follow the setup from ~\cite{vera}, and fine-tune on a subset comprising 10 samples from each class.

\begin{table*}[h]
\centering
\small
\addtolength{\tabcolsep}{-4.8pt}
\caption{Performance (Accuracy) on Mathematical Reasoning (GSM-8K and MATH). \#Params denote the number of trainable parameters. \textbf{bold} and \underline{underline} represent best and second best performing PEFT method, respectively. \textsc{SVFT} offers superior/competitive performance at much lower \#Params. For \(\textsc{SVFT}^R_d\), we set \(d=16\) for Gemma and \(d=12\) for LLaMA-3 models.}
\begin{tabular}{@{}lccccccccc@{}}
\toprule
\multirow{2}{*}{{\textbf{Method}}} & \multicolumn{3}{c}{Gemma-2B} & \multicolumn{3}{c}{Gemma-7B} & \multicolumn{3}{c}{LLaMA-3-8B} \\ \cmidrule(l){2-4} \cmidrule(l){5-7} \cmidrule(l){8-10} 
 & \textbf{\#Params} & \textbf{GSM-8K} & \textbf{MATH} & \textbf{\#Params} & \textbf{GSM-8K} & \textbf{MATH} & \textbf{\#Params} & \textbf{GSM-8K} & \textbf{MATH} \\ \midrule 
 \midrule
 Full-FT & 2.5B & 52.69 & 17.94 & 8.5B & 74.67 & 25.70 & 8.0B & 64.13 & 16.24 \\ \midrule
\(\text{LoRA}_{r=32}\) & 26.2M & 43.06 & 15.50 & 68.8M & \underline{76.57} & 29.34 & 56.6M & \textbf{75.89} & \textbf{24.74} \\
\(\text{DoRA}_{r=16}\) & 13.5M & \underline{44.27} & \textbf{16.18} & 35.5M & 74.52 & \underline{29.84} & 29.1M & 75.66 & \textbf{24.72} \\ \midrule
\(\text{BOFT}^{b=8}_{m=2}\) & 1.22M & 36.01 & 12.13 & 2.90M & 71.79 & 28.98 & 4.35M & 67.09 & 21.64 \\
\(\text{DoRA}_{r=1}\) & 1.19M & 35.25 & 13.04 & 3.26M & 74.37 & 26.28 & 2.55M & 68.30 & 21.96 \\
\(\text{LoRA}_{r=1}\) & 0.82M & 32.97 & 13.04 & 0.82M & 72.4 & 26.28 & 1.77M & 68.84 & 20.94 \\
\(\text{VeRA}_{r=1024}\) & 0.63M & 36.77 & 14.12 & 0.43M & 71.11 & 27.04 & 0.98M & 63.76 & 20.28 \\
\rowcol \(\textsc{SVFT}^P\) & 0.19M & 40.34 & 14.38 & 0.43M & 73.50 & 27.30 & 0.48M & \underline{69.22} & 20.44 \\
\rowcol \(\textsc{SVFT}^R_d\) & 6.35M & \textbf{50.03} & \underline{15.56} & 19.8M & \textbf{76.81} & \textbf{29.98} & 13.1M & \textbf{75.90} & \underline{24.22} \\ \bottomrule
\end{tabular}
\label{tab:math_reasoning}
\end{table*}

\begin{table*}[ht]
\centering
\small
\addtolength{\tabcolsep}{-1.5pt}
\caption{Evaluation results on eight commonsense reasoning benchmarks with Gemma-7B. We follow~\cite{dora} for hyperparameter configurations, and report accuracy for all tasks. HS and WG denote HellaSwag~\cite{hellaswag} and WinoGrande~\cite{winogrande}, respectively. $\text{SVFT}^P$ offers competitive performance at a fraction of \#Params. $\text{SVFT}_{d=8}^{B}$ can match $\text{LoRA}_{r=32}$  with $\sim$7x fewer parameters.}
\begin{tabular}{lcccccccccc}
\toprule
\textbf{Method} & \textbf{\#Params} & \textbf{BoolQ} & \textbf{PIQA} & \textbf{SIQA} & \textbf{HS} & \textbf{WG} & \textbf{ARC-e} & \textbf{ARC-c} & \textbf{OBQA} & \textbf{Average} \\ \midrule \midrule
Full-FT & 8.5B & 72.32 & 87.32 & 76.86 & 91.07 & 81.76 & 92.46 & 82.76 & 89.00 & 84.19 \\ \midrule
\(\text{LoRA}_{r=32}\) & 68.8M & \underline{71.55} & \textbf{87.95} & \textbf{77.27} & \underline{91.80}  & \textbf{79.71} & 92.67 & 82.16 & \textbf{86.40}  & \textbf{83.69}  \\
\(\text{DoRA}_{r=16}\) & 35.5M  & 71.46 & \underline{87.59} & \underline{76.35} & \textbf{92.11} & 78.29 & 92.00 & 80.63 & 85.60 & 83.00 \\ \midrule
\(\text{DoRA}_{r=1}\) & 3.31M  & 68.22 & 86.72 & 75.23 & 91.14 & 78.13 & 91.87 & \textbf{83.19} & \underline{86.20} & 82.59 \\
\(\text{VeRA}_{r=2048}\) & 1.49M & 64.25 & 86.28 & 74.04 & 86.96  & 69.00  & \underline{92.76} & 82.33 & 82.00  & 79.70 \\
\(\text{LoRA}_{r=1}\) & 0.82M  & 65.44 & 86.28 & 75.02 & 89.91 & 75.92 & 91.79 & 81.91 & 85.40 & 81.46 \\
\rowcol \(\textsc{SVFT}^P\) & 0.51M & 67.92 & 86.45 & 75.47 & 86.92 & 74.03 & 91.80 & 81.23 & 83.00 & 80.85 \\
\rowcol \(\textsc{SVFT}^B_{d=8}\) & 9.80M  & \textbf{71.90} & 86.98 & 76.28 & 91.55 & \underline{78.76} & \textbf{92.80} & \underline{83.11} & 85.40 & \underline{83.35} \\ \bottomrule
\end{tabular}
\label{tab:commonsense_resoning}
\end{table*}

\begin{table*}[t]
\centering
\small
\addtolength{\tabcolsep}{-4.2pt}
\caption{DeBERTaV3\textsubscript{base} with different adaptation methods on the GLUE benchmark. We report matched accuracy for MNLI, Matthew’s correlation for CoLA, Pearson correlation for STS-B, and accuracy for other tasks. Higher is better for all tasks. * indicates numbers published in prior work. }
\begin{tabular}{lcccccccccc}\toprule
\textbf{Method} & \textbf{\#Params} & \textbf{MNLI} & \textbf{SST-2} & \textbf{MRPC} & \textbf{CoLA} & \textbf{QNLI} & \textbf{QQP} & \textbf{RTE} & \textbf{STS-B} & \textbf{Avg.}\\ \midrule \midrule
Full-FT* & 184M & 89.90 & 95.63 & 89.46 & 69.19 & 94.03 & \textbf{92.40} & 83.75 & 91.60 & 88.25 \\ \midrule
\( \text{LoRA*}_{r=8} \) & 1.33M & \textbf{90.65} & 94.95 & \underline{89.95} & 69.82 & 93.87 & 91.99 & 85.20 & 91.60 & 88.50 \\
\( \text{DoRA}_{r=4} \) & 0.75M & 89.92 & 95.41 & 89.10 & 69.37 & 94.14 & 91.53 & 87.00 & \underline{91.80} & 88.53 \\
\( \text{BOFT*}^{b=8}_{m=2} \) &  0.75M & \underline{90.25} & \textbf{96.44} & \textbf{92.40} & \textbf{72.95} & \underline{94.23} & \underline{92.10} & \textbf{88.81} & \textbf{91.92} & \textbf{89.89} \\ 
\midrule
\( \text{LoRA}_{r=1} \) & 0.17M & 90.12 & 95.64 & 86.43 & 69.13 & 94.18 & 91.43 & 87.36 & 91.52 & 88.23 \\
\( \text{VeRA}_{r=1024} \) & 0.09M & 89.93 & 95.53 & 87.94 & 69.06 & 93.24 & 90.4 & 87.00 & 88.71 & 87.73 \\
\rowcol \(\textsc{SVFT}^P\) & 0.06M & 89.69 & 95.41 & 88.77 & 70.95 & \textbf{94.27} & 90.16 & 87.24 & \underline{91.80} & 88.54  \\
\rowcol \textsc{SVFT$^R_{d=2}$} & 0.28M & 89.97 & \underline{95.99} & 88.99 & \underline{72.61} & 93.90 & 91.50 & \underline{88.09} & 91.73 & \underline{89.10} \\
\bottomrule
\end{tabular}
\label{tab:glue_deberta_results}
\end{table*}

\begin{table*}[t]
\centering
\small
\addtolength{\tabcolsep}{1pt}
\caption{Performance on image classification benchmarks. For LoRA, DoRA and \(\textsc{SVFT}^B\), we adapt \{Q, K, V, U, D\} modules of the transformer. 
For \(\textsc{SVFT}^P\), we adapt only \{Q, V\} to keep it comparable with VeRA. 
We report accuracy for all tasks.}
\begin{tabular}{lcccccc}\toprule
\multirow{2}{*}{\textbf{Method}} & \multicolumn{3}{c}{ViT-B} & \multicolumn{3}{c}{ViT-L} \\ \cmidrule(lr){2-4} \cmidrule(lr){5-7}
    &  \textbf{\#Params} & \textbf{CIFAR100} & \textbf{Flowers102} & \textbf{\#Params} &  \textbf{Food101} & \textbf{Resisc45} \\ \midrule \midrule
Head & - & 78.25 & 98.42 & - & 75.57 & 64.10 \\
Full-FT & 85.8M & 85.35 & 98.37 & 303.3M  & 77.83 & 76.83 \\ \midrule
\( \text{LoRA}_{r=8} \) & 1.32M & 84.10 & \underline{99.23} & 3.54M & 77.13 & \textbf{79.62} \\
\( \text{DoRA}_{r=8} \) & 1.41M & 85.03 & \textbf{99.30} & 3.76M & 76.41 & \underline{78.32} \\ \midrule
\( \text{BOFT}^{b=4}_{m=4} \) & 0.11M & \underline{85.54} & 98.59 & 2.95M & \textbf{78.42} & 74.70 \\
\( \text{LoRA}_{r=1} \) & 0.16M & 84.86 & 96.88 & 0.44M & 75.97 & 78.02 \\
\( \text{DoRA}_{r=1} \) & 0.25M & 84.46 & 99.15 & 0.66M & 75.90 & 78.02 \\
\(\text{VeRA}_{r=256} \) & 24.6K & 83.38 & 98.59 & 0.06M & 75.97 & 72.44 \\ \midrule
\rowcol $\textsc{SVFT}^P$ & 18.5K & 83.85  & 98.93  & 0.05M & 75.95  & 71.97  \\
\rowcol \( \textsc{SVFT}^B_{d=2} \) & 0.27M & 84.72  & \textbf{99.28}  & 0.74M  & \underline{77.94}  & \textbf{79.70}  \\
\rowcol \( \textsc{SVFT}^B_{d=8} \) & 0.93M & \textbf{85.69}  & 98.88  & 2.5M  & \textbf{78.36}  & 73.83  \\
\bottomrule
\end{tabular}
\vspace{0.2cm}
\label{tab:vit_results}
\end{table*}

\section{Results}
\subsection{Performance on Language Tasks}
\paragraph{Natural Language Generation.} We present results on mathematical question answering against baseline PEFT techniques across three base models -- varying from 2B to 8B parameters in \autoref{tab:math_reasoning}. To ensure a comprehensive comparison, we test baseline techniques (LoRA, DoRA) with different configurations, and varying hyper-parameters like rank to cover a range of learnable parameters from low to high. Note that even when the rank is as low as 1, both methods yield more trainable parameters than \(\textsc{SVFT}^P\). \(\textsc{SVFT}^P\) (\(\sim\)0.2M) shows as much as \(18\%\) relative improvement over techniques that use 6\(\times\) more trainable parameters (\(\text{BOFT}^{b=8}_{m=2}\), \(\text{LoRA}_{r=1}\)). Against techniques of comparable sizes (VeRA), \(\textsc{SVFT}^P\) achieves \textbf{15.5\%} relative improvement on average. Even in the default regime, \(\textsc{SVFT}^{R}_{d}\) matches techniques with at least \(3\times\) more trainable parameters. Notably, on GSM-8K, \(\textsc{SVFT}^{R}_{d}\) again achieves \textbf{96\%} of the full fine-tuning performance, while \(\text{DoRA}_{r=16}\) recovers 86\% with \(2\times\) more parameters than \(\textsc{SVFT}^{R}_{d}\). 

\paragraph{Commonsense Reasoning.} In \autoref{tab:commonsense_resoning}, we compare performance on commonsense reasoning benchmarks with Gemma-7B, and observe similar trends. In the lower and moderately parameterized regime (\(\sim\)0.43M), \(\textsc{SVFT}^{P}\) shows competitive performance in comparison to \(\textsc{LoRA}_{r=1}\) and \(\text{DoRA}_{r=1}\), which have 1.9\(\times\) and 7.7\(\times\) more parameters, respectively. Against VeRA, which trains 3.5\(\times\) more parameters, \(\textsc{SVFT}^{P}\) shows a relative improvement of \(\sim\)\textbf{1.16}\%. Similarly, \(\textsc{SVFT}^{B}_{d=8}\) also matches or exceeds methods that use up to 7\(\times\) more trainable parameters. For instance, \(\textsc{SVFT}^{B}_{d=8}\) attains an average performance of 83.35\% with only 9.8M parameters, closely matching \(\text{LoRA}_{r=16}\) (83.69\%, 68.8M parameters). We observe similar trends with Gemma-2B (refer \autoref{tab:commonsese_gemma2B}).

\paragraph{Natural Language Understanding.} Results on the GLUE benchmark are summarized in \autoref{tab:glue_deberta_results}. \textsc{SVFT} matches \(\text{LoRA}_{r=8}\) and \(\text{DoRA}_{r=4}\) which use \textbf{12-22}\(\times\) more trainable parameters. Similarly, when compared to OFT and BOFT, \(\textsc{SVFT}^{P}\) maintains a comparable average performance despite being 12\(\times\) smaller. These results highlight \textsc{SVFT}'s ability to strike a balance between parameter efficiency and performance, making it an attractive PEFT choice for simple classification tasks. 

\textbf{Parameter efficiency.} In \autoref{fig:nlg_plots}, we plot the performance of SVFT on mathematical reasoning and commonsense reasoning against other PEFT techniques across a range of configurations. Across trainable parameter budgets ranging from lowest to highest, \textsc{SVFT} obtains the best overall performance, matching methods that require significantly more trainable parameters. These results establish \textsc{SVFT} as a Pareto-dominant approach for parameter-efficient fine-tuning.

\subsection{Performance on Vision Tasks}

\begin{wrapfigure}{R}{0.4\textwidth}
\centering
\includegraphics[width=0.4\textwidth]{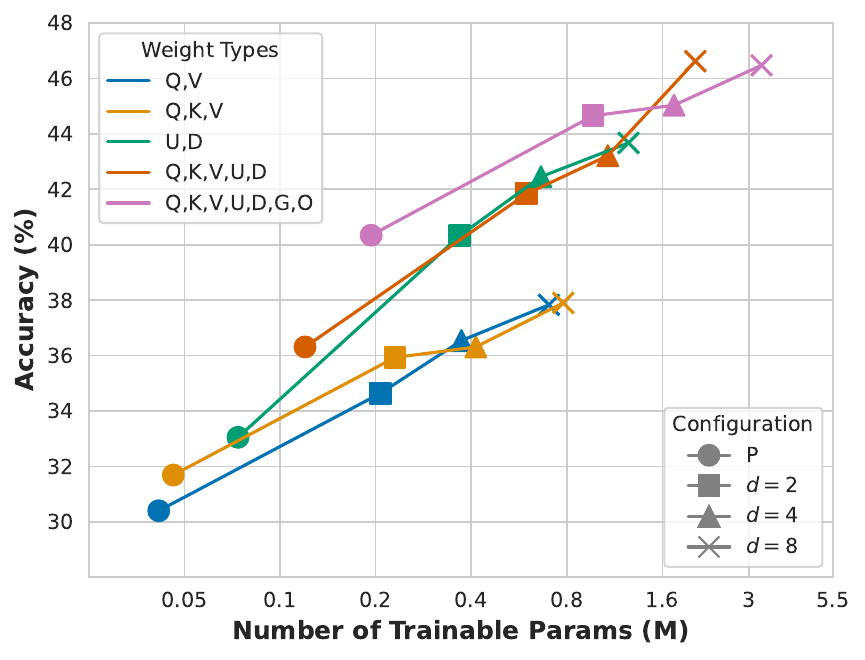}
\caption{Performance variation with \(\textsc{SVFT}^{B}_{d}\) based on the adapted weight matrices -- GSM-8K with Gemma-2B. Adapting more target weight types results in greater gains in performance. Interestingly, for a fixed parameter budget, adapting $\mU$ and $\mD$ weight types gives greater lifts than adapting $\mQ$ and $\mV$.}
\label{fig:ablation_target}
\vspace{-1.3cm}
\end{wrapfigure}

\autoref{tab:vit_results} contrasts \textsc{SVFT} against other PEFT techniques on image classification benchmarks using ViT-B and ViT-L models. 
For ViT-B, \(\textsc{SVFT}^B_{d=8}\) surpasses full fine-tuning performance along with \(\text{LoRA}_{r=8}\) and \(\text{DoRA}_{r=8}\) on CIFAR-100. \(\textsc{SVFT}^B_{d=2}\) matches \(\text{LoRA}_{r=8}\) and \(\text{DoRA}_{r=8}\) on Flowers102 with up to \(5\times\) fewer parameters. 
For ViT-L, \(\textsc{SVFT}^{B}_{d}\) also demonstrates superior or competitive performance on both Food101 and Resisc45, with significantly lower trainable parameters compared to both fully fine-tuned models and other state-of-the-art PEFT approaches.

\subsection{Contribution of Each Weight Type}

In \autoref{fig:ablation_target}, we investigate the contribution of each weight type. Starting with the base configuration, we apply \(\textsc{SVFT}^{B}_{d}\) to the \(\mQ\) and \(\mV\) weights in each transformer block and report the performance. We then incrementally add the remaining weight modules (\(\mK, \mU, \mD, \mO, \mG\)) and observe the changes in performance. For each configuration, we also vary the trainable parameters by incrementing the total learnable off-diagonals.

Note that applying \(\textsc{SVFT}^{B}_{d}\) to \(\mU, \mD, \mO,\) and \(\mG\) does not increase trainable parameters as much as applying LoRA/DoRA to these modules (\autoref{tab:complexity_analysis}). For example, for a large matrix of shape \(d_1 \times d_2\), \(\text{LoRA}_{r=1}\) learns \(d_1 + d_2\) parameters, while \(\textsc{SVFT}^{P}\) learns \(\min(d_1, d_2)\) parameters. We observe that adapting only \(\mU\) and \(\mD\) with \textsc{SVFT} yields up to a \(10\%\) relative improvement over adapting \(\mQ\) and \(\mV\) for the same parameter budget (\(\sim0.8M\)). Our findings indicate that adapting more weight types enhances performance.\looseness=-1

\begin{table*}[h]
\begin{minipage}{0.47\columnwidth}
    \centering
    \caption{Results on fine-tuning Gemma-2B with \textsc{SVFT} using different $\mM$ parameterizations.}
    \addtolength{\tabcolsep}{-2pt}
    \begin{tabular}{@{}l@{\hspace{0.7\tabcolsep}}ccc@{}}
    \toprule
    \textbf{Structure} & \textbf{\#Params}  & \textbf{GSM-8K} & \textbf{MATH} \\ \midrule \midrule
    Plain & 0.2M & 40.34 & 14.38 \\ \midrule
    \multirow{2}{*}{Banded} & 3.3M & 46.47 & \textbf{16.04} \\
     & 6.4M & 47.84 & 15.68 \\ \midrule
    \multirow{2}{*}{Random} & 3.3M & 47.76 & \underline{15.98} \\
     & 6.4M & \textbf{50.03} & 15.56 \\ \midrule
    \multirow{2}{*}{Top-$k$} & 3.3M & 48.00 & 15.80 \\
     & 6.4M & \underline{49.65} & 15.32 \\ \bottomrule
    \end{tabular}
    \label{table:abl_struc}
\end{minipage}\hfill
\begin{minipage}{0.48\columnwidth}
    \centering
    \addtolength{\tabcolsep}{-2pt}
    \caption{Impact of pre-trained weight quality. Results on GSM-8K after fine-tuning on Pythia-2.8B checkpoints at different stages of pre-training (PT).
    Compared to LoRA, \textsc{SVFT} benefits more from better pre-trained weights. \textsc{SVFT} outperforms LoRA in both cases.
    }
    \begin{tabular}{@{}lcccc@{}}
    \toprule
    \multirow{2}{*}{\textbf{Method}} & \multirow{2}{*}{\textbf{\#Params}} & \multicolumn{2}{c}{PT Steps} & \multicolumn{1}{c}{\multirow{2}{*}{\textbf{$\Delta$Perf}}} \\ \cmidrule(lr){3-4}
     &  & \multicolumn{1}{c}{\textbf{39K}} & \multicolumn{1}{c}{\textbf{143K}} & \multicolumn{1}{c}{} \\ \midrule \midrule
    Full-FT & 2.5B & 21.00 & 30.09 & 9.09 \\
    LoRA & 5.24M & 11.22 & 18.95 & 7.73 \\
    \rowcol \textsc{SVFT} & 5.56M & 15.08 & 23.19 & 8.11 \\ \bottomrule
    \end{tabular}
    \label{table:abl_pythia}
  \end{minipage}
\end{table*}

\subsection{Impact of \({M}\)'s Structure on Performance}

We analyze the impact of different parameterizations of \(\mM\) (Plain, Banded, Random, Top-\(k\)) on downstream performance. To ensure a fair comparison, we match the number of trainable coefficients across all variants. As shown in Table \ref{table:abl_struc}, both Random and Top-\(k\) variants outperform Banded on the GSM-8K dataset. However, this improvement comes at the cost of performance on MATH. This observation suggests that the choice of parameterization has a significant impact on model performance, and the effectiveness of a particular structure may vary depending on the downstream task.\looseness=-1

\subsection{Impact of Pre-trained Weight Quality}
A key feature of \textsc{SVFT} is that the weight update depends on the pre-trained weights \(\mW\). We therefore ask the following question: \textit{Does the quality of pre-trained weights have a disproportionate impact on \textsc{SVFT}?} To answer this, we consider two checkpoints from the Pythia suite~\cite{pythia} at different stages of training, i.e., 39K steps and 143K steps, respectively. We fine-tune each of these checkpoints independently with Full-FT, LoRA, and \textsc{SVFT}. We then compare the increase in performance (\(\Delta\)Perf). As shown in \autoref{table:abl_pythia}, compared to LoRA, \textsc{SVFT} benefits more from better pre-trained weights. We also note that \textsc{SVFT} outperforms LoRA in both settings, suggesting that the benefits of inducing a \(\Delta\mW\) that explicitly depends on \(\mW\) are beneficial even when \(\mW\) is sub-optimal.

\section{Discussion}
\label{sec:limitations}

\textbf{Limitations.} Despite significantly reducing learnable parameters and boosting performance, \textsc{SVFT} incurs some additional GPU memory usage. 
Unlike LoRA and its variants, \textsc{SVFT} necessitates computing the SVD and storing both left and right singular vectors. 
While memory consumption remains lower than BOFT, it's roughly double that of LoRA. 
We mitigate this in our work by employing system-level optimizations like mixed-precision weights (e.g., bfloat16). 
However, similar to the scaling explored in~\cite{flora}, memory usage should amortize with the increasing scale of adaptation tasks. 
In future work we will explore quantization and other techniques to address memory concerns.

\textbf{Broader Impact.} 
Our work enables easier personalization of foundational models, which can have both positive and negative societal impacts. Since our method provides computational efficiency (smaller parameter footprint), it will be less expensive to enable personalization.

\newpage
\section{Conclusion}
\label{sec:conclusion}
This work introduces \textsc{SVFT}, a novel and efficient PEFT approach that leverages the structure of pre-trained weights to determine weight update perturbations. We propose four simple yet effective sparse parameterization patterns, offering flexibility in controlling the model's expressivity and the number of learnable parameters. Extensive experiments on language and vision tasks demonstrate \textsc{SVFT}'s effectiveness as a PEFT method across diverse parameter budgets. Furthermore, we theoretically show that \textsc{SVFT} can induce higher-rank perturbation updates compared to existing methods, for a fixed parameter budget. In future work, we aim to develop principled methods to generate sparsity patterns, potentially leading to further performance improvements.

\section*{Acknowledgements}
\label{sec:acknowledgements}
We thank CISPA Helmholtz Center for Information Security and Greg Kuhlmann for their invaluable support in facilitating this research. We also appreciate Anubhav Goel for his helpful discussions and support.


\bibliographystyle{plain}
\bibliography{bibliography}

\newpage

\appendix
\section*{Appendix}
The appendix is organized as follows.
\begin{itemize}
    \item In ~\autoref{app:proofs}, we give proofs for the lemmas outlined in ~\ref{subsection:insights}. 
    \item In ~\autoref{app:param_count_analysis}, we compare how the trainable parameters count for different PEFT techniques (LoRA, DoRA, VeRA) versus our method \textsc{SVFT}.
    \item In ~\autoref{app:additional_exp}, we describe results for additional experiments and provide implementation details for all the experiments.
\end{itemize}

\section{Proofs}
\label{app:proofs}
We provide brief proofs for the \textit{Structure}, \textit{Expressivity} and the \textit{Rank} lemmas for SVFT:

\begin{enumerate}
\item[{\em (a)}] {\em Structure:} 
If \(\mM\) is diagonal, then the final matrix \(\mW_0 + U \mM V^T\) can be written as \\ 
\(U (\Sigma + \mM) V^T\) since \(\mW_0 = U \Sigma V^T\), where \((\Sigma + \mM)\) is also a diagonal matrix. 
Thus, \(U (\Sigma + \mM) V^T\) is a valid and unique SVD of \(\mW_0 + U \mM V^T\) up to sign flips in the singular vectors.

\item[{\em (b)}] {\em Expressivity:} Finding \(\mM\) for any target matrix \(P\) of size \(d_1 \times d_2\) such that \(P = \mW_0 + U \mM V^T\) is the same as finding \(\mM\) for a new target matrix \(P' = P - \mW_0\) such that \(P' = U \mM V^T\).
For a full SVD, the dimension of \(\mM\) is \(d_1 \times d_2\) and since the dimension of \(P'\) is also \(d_1 \times d_2\), \(P' = U \mM V^T\) is a bijection and \(\mM = U^T (P - \mW_0) V\) (since \(U\) and \(V\) are orthogonal).

\item[{\em (c)}]{\em Rank:} If \(\mM\) has \(k\) non-zero elements, then the rank of the update \(U \mM V^T\) will be upper bounded by \(k\) (since by Gaussian elimination, \(k\) or less elements will remain, the best case being all \(k\) elements in the diagonal).
We also know that the rank is upper bounded by \(\min\{d_1, d_2\}\), giving an achievable upper bound on the rank as \(\min\{k, \min\{d_1, d_2\}\}\).
\end{enumerate}

\section{Parameter Count Analysis}
\label{app:param_count_analysis}
\begin{table*}[h]
\centering
\small
\caption{Parameter count analysis. $L_{\text{tuned}}$, $D_{\text{model}}$, $r$, $k$ denote total layers being adapted, hidden dimension, rank, and additional off-diagonals respectively. }
\begin{tabular}{lc}
\toprule
\textbf{Method} & \textbf{Trainable Parameter Count} \\ \midrule
LoRA & $2\times L_{\text{tuned}} \times D_{\text{model}} \times r $ \\
DoRA &  $L_{\text{tuned}} \times D_{\text{model}}\times (2r + 1)$\\
VeRA &  $L_{\text{tuned}} \times (D_{\text{model}} + r)$\\ 
\rowcol \(\text{SVFT}^P\) &  $L_{\text{tuned}} \times D_{\text{model}}$\\ 
\rowcol \(\text{SVFT}^B_{d=k}\) &  $L_{\text{tuned}}\times(D_{\text{model}} \times k + (D_{\text{model}}-k)(k+1))$\\ \bottomrule 
\end{tabular}
\label{tab:complexity_analysis}
\end{table*}

\section{Additional Experiments and Implementation Details}
\label{app:additional_exp}

All of our experiments are conducted on a Linux machine (Debian GNU) with the following specifications: 2xA100 80 GB, Intel Xeon CPU @ 2.20GHz with 12 cores, and 192 GB RAM. For all our experiments (including baseline experiments), we utilize hardware-level optimizations like mixed weight precision (e.g., bfloat16) whenever possible.

\subsection{Commonsense Reasoning Gemma-2B}
We evaluate and compare SVFT variants against baseline PEFT methods on commonsense reasoning tasks with Gemma-2B model and tabulate results in~\autoref{tab:commonsese_gemma2B}.

\begin{table*}[ht]
\centering
\small
\addtolength{\tabcolsep}{-4.5pt}
\caption{Results with Gemma-2B on eight commonsense reasoning benchmarks. We follow~\cite{dora} for hyperparameter configurations, and report accuracy for all tasks.}
\begin{tabular}{lcccccccccc}
\toprule
\textbf{Method} & \textbf{\#Params} & \textbf{BOOLQ} & \textbf{PIQA} & \textbf{SIQA} & \textbf{HellaSwag} & \textbf{Winogrande} & \textbf{ARC-E} & \textbf{ARC-C} & \textbf{OBQA} & \textbf{Average} \\ \midrule \midrule
 Full-FT & 2.5B & 63.57 & 74.1 & 65.86 & 70.00 & 61.95 & 75.36 & 59.72 & 69 & 67.45 \\
\(\text{LoRA}_{r=32}\) & 26.2M & 63.11 & 73.44 & 63.20 & 47.79 & 52.95 & 74.78 & 57.16 & 67.00 & 62.43 \\
\(\text{LoRA}_{r=16}\) & 13.5M & 62.87 & 73.93 & 65.34 & 53.16 & 55.51 & 76.43 & 59.55 & 68.4 & 64.40 \\
\(\text{BOFT}_{m=2}^{b=8}\) & 1.22M & 59.23 & 63.65 & 47.90 & 29.93 & 50.35 & 59.04 & 42.66 & 41.00 & 49.22 \\
\(\text{VeRA}_{r=2048}\) & 0.66M & 62.11 & 64.31 & 49.18 & 32.00 & 50.74 & 58.08 & 42.83 & 42.6 & 50.23 \\
\(\text{LoRA}_{r=1}\) & 0.82M & 62.2 & 69.31 & 56.24 & 32.47 & 51.53 & 69.52 & 48.8 & 56.4 & 55.81 \\
\(\text{DoRA}_{r=1}\) & 1.19M & 62.17 & 68.77 & 55.93 & 32.95 & 51.22 & 68.81 & 48.72 & 55.6 & 55.52 \\
\rowcol \(\textsc{SVFT}^P\) & 0.19M & 62.26 & 70.18 & 56.7 & 32.47 & 47.04 & 69.31 & 50.08 & 58.4 & 55.81 \\
\rowcol \(\textsc{SVFT}^B_{d=16}\) & 6.35M & 63.42 & 73.72 & 63.86 & 71.21 & 59.58 & 73.69 & 54.77 & 66.6 & 65.86 \\ \bottomrule
\end{tabular}
\label{tab:commonsese_gemma2B}
\end{table*}

\subsection{Additional Vision Experiments}

For vision tasks, we compare the \textsc{SVFT} banded variants and \textsc{SVFT} plain with baseline PEFT methods on classification vision tasks using ViT-Base and ViT-Large models in ~\autoref{tab:app_vit_results}.

\begin{table*}[t]
\centering
\small
\addtolength{\tabcolsep}{-4.2pt}
\caption{Performance on image classification benchmarks. For LoRA, DoRA and \(\textsc{SVFT}^{B}_{d}\), we adapt \{Q, K, V, U, D\} modules of the transformer. 
For \(\textsc{SVFT}^P\), we adapt only \{Q, V\} to keep it comparable with VeRA. 
We report accuracy for all tasks.}
\resizebox{0.98\textwidth}{!}{
\begin{tabular}{lcccccccccc}\toprule
\multirow{2}{*}{\textbf{Method}} & \multicolumn{5}{c}{ViT-B} & \multicolumn{5}{c}{ViT-L} \\ \cmidrule(lr){2-6} \cmidrule(lr){7-11}
    &  \textbf{\#Params} & \textbf{CIFAR100} & \textbf{Flowers102} & \textbf{Food101} & \textbf{Resisc45} & \textbf{\#Params} & \textbf{CIFAR100} & \textbf{Flowers102} & \textbf{Food101} & \textbf{Resisc45} \\ \midrule \midrule
Head & - & 78.25 & 98.42 & 74.93 & 59.95 & - & 82.95 & 98.75 & 75.57 & 64.10 \\
Full-FT & 85.8M & 85.35 & 98.37 & 76.32 & 68.03 & 303.3M & 86.56 & 97.87 & 77.83 & 76.83 \\
\( \text{LoRA}_{r=8}\) & 1.32M & 84.41 & 99.23 & 76.02 & 76.86 & 0.35M & 86.00 & 97.93 & 77.13 & 79.62 \\
\( \text{DoRA}_{r=8}\) & 1.41M & 85.03 & 99.30 & 75.88 & 76.95 & 3.76M & 83.55 & 98.00 & 76.41 & 78.32 \\
\( \text{BOFT}^{b=2}_{m=2} \) & 0.07M & 85.55 & 98.54 & 76.06 & 67.70 & 0.20M & 87.84 & 97.95 & 77.90 & 73.97 \\
\( \text{BOFT}^{b=4}_{m=4} \) & 0.11M & 85.54 & 98.59 & 76.51 & 69.44 & 0.30M & 87.72 & 97.95 & 78.42 & 74.70 \\
\midrule
\( \text{LoRA}_{r=1}\) & 0.16M & 84.86 & 96.88 & 73.35 & 76.33 & 0.44M & 85.97 & 98.28 & 75.97 & 78.02 \\
\( \text{DoRA}_{r=1}\) & 0.25M & 84.46 & 99.15 & 74.80 & 77.06 & 0.66M & 84.06 & 98.11 & 75.90 & 78.02 \\
VeRA & 24.6K & 83.38 & 98.59 & 75.99 & 70.43 & 61.4K & 86.77 & 98.94 & 75.97 & 72.44 \\ \midrule
\rowcol \(\textsc{SVFT}^P\) & 18.5K & 83.85  & 98.93  & 75.68  & 67.19  & 49.2K & 86.74  & 97.56  & 75.95  & 71.97  \\
\rowcol \( \textsc{SVFT}^B_{d=2}\) & 0.28M & 84.72  & 99.28  & 75.64  & 72.49  & 0.74M & 86.59  & 98.24  & 77.94  & 79.70  \\
\rowcol \( \textsc{SVFT}^B_{d=4}\) & 0.50M & 83.17  & 98.52  & 76.54  & 66.65  & 1.32M & 87.10  & 97.71  & 76.67  & 71.10  \\
\rowcol \( \textsc{SVFT}^B_{d=8}\) & 0.94M & 85.69  & 98.88  & 76.70  & 70.41  & 2.50M & 87.26  & 97.89  & 78.36  & 73.83  \\
\bottomrule
\end{tabular}
}
\vspace{0.2cm}
\label{tab:app_vit_results}
\end{table*}

\subsection{Are All Singular Vectors Important?}
To determine the importance of considering all singular vectors and singular values during fine-tuning, we reduce the rank of \(\mU\) and \(\mV\), and truncate \(\mSigma\) and \(\mM\) to an effective rank of \(r\). If the original weight matrix \(\mW \in \mathbb{R}^{m\times n}\), then after truncation, \(\mU \in \mathbb{R}^{m\times r}, \mV \in \mathbb{R}^{n\times r}\). This truncation significantly reduces the number of trainable parameters, so we compensate by increasing the number of off-diagonal coefficients (\(d\)) in \(\mM\). 

Our results, with four different configurations of \(r\) and \(d\), are presented in \autoref{table:abl_trunc}. The findings show that a very low rank (\(r=128\)) leads to poor performance, even when parameters are matched. A reasonably high rank of \(r=1536\), which is 75\% of the full rank, still fails to match the performance of the full-rank variant that has 0.25\(\times\) the trainable parameters. This indicates that all singular vectors significantly contribute to the end task performance when fine-tuning with \textsc{SVFT}, and that important information is lost even when truncating sparingly. 

\begin{table}[ht]
\centering
\small
\addtolength{\tabcolsep}{-2pt}
\caption{Performance with varying rank ($r$) and the off-diagonal elements ($d$) of $\mM$. When $r=2048$, the update is full-rank.}
\begin{tabular}{ccccc}
\toprule
\textbf{Rank ($r$)} & \textbf{Diags ($d$)} & \textbf{\#Params} & \textbf{GSM-8K} & \textbf{MATH} \\ \midrule \midrule
128 & 64 & 1.55M & 0.98 & 0.21 \\
1536 & - & 0.15M & 16.37 & 3.64 \\
1536 & 2 & 0.74M & 25.01 & 6.04 \\
\rowcol 2048 & - & 0.19M & \textbf{40.34} & \textbf{14.38} \\ \bottomrule
\end{tabular}
\label{table:abl_trunc}
\end{table}

\subsection{Performance vs Total Trainable Parameters}

\begin{figure*}[ht]
\centering
    \includegraphics[width=0.95\linewidth]{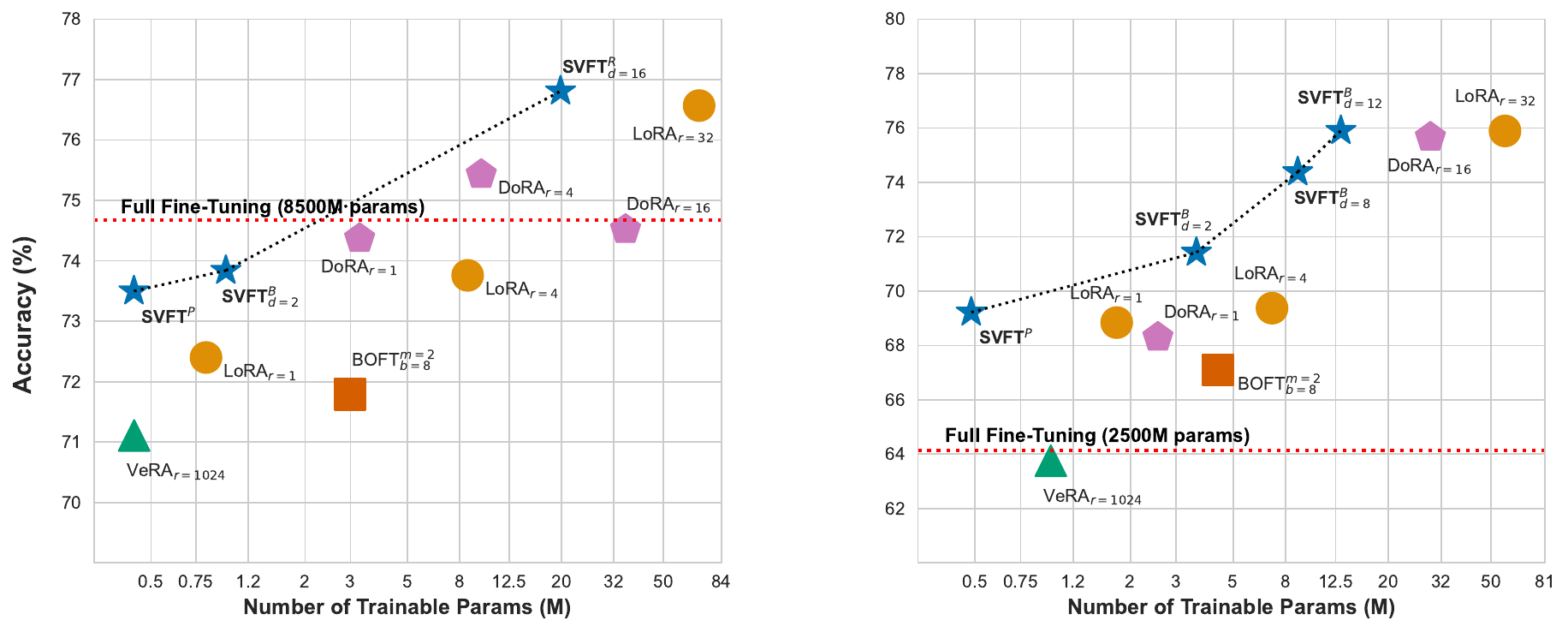}
    \caption{Performance versus total trainable parameters for GSM-8K on Gemma-7B (left) and LLaMA-3-8B (right).}
    \label{fig:appendix_pareto_plot}
\end{figure*}

In addition to the experiments performed in \autoref{fig:nlg_plots} for Gemma-2B on challenging natural language generation (NLG) tasks like GSM-8K and Commonsense Reasoning, we also plot the performance vs total trainable parameters for larger state-of-the-art models like Gemma-7B and LLaMA-3-8B on GSM-8K. \autoref{fig:appendix_pareto_plot} further demonstrates SVFT's Pereto-dominance. On larger models, we observe that full-finetuning overfits, leading to sub-optimal performance in comparison to PEFT methods.

\subsection{Settings for Language Tasks}
\label{app:language}
\paragraph{Natural Language Understanding.} We fine-tune the DeBERTaV3\textsubscript{base}~\cite{debertav3} model and apply \textsc{SVFT} to all linear layers in every transformer block of the model. We only moderately tune the batch size, learning rate, and number of training epochs. We use the same model sequence lengths used by ~\cite{boft} to keep our comparisons fair. The hyperparameters used in our experiments can be found in ~\autoref{tab:glue_deberta_hparams}.  
\begin{table*}[h]
\centering
\caption{Hyperparameter setup used for DeBERTaV3\textsubscript{base} on the GLUE benchmark.}
\resizebox{0.99\textwidth}{!}{
\begin{tabular}{llcccccccc}\toprule
\textbf{Method} & \textbf{Dataset} & \textbf{MNLI} & \textbf{SST-2} & \textbf{MRPC} & \textbf{CoLA} & \textbf{QNLI} & \textbf{QQP} & \textbf{RTE} & \textbf{STS-B}\\ \midrule
& Optimizer & \multicolumn{ 8}{c}{AdamW} \\
& Warmup Ratio & \multicolumn{ 8}{c}{0.1} \\
& LR Schedule & \multicolumn{ 8}{c}{Linear} \\ 
& Learning Rate (Head) & \multicolumn{ 8}{c}{6E-03} \\ 
& Max Seq. Len. & 256 & 128 & 320 & 64 & 512 & 320 & 320 & 128 \\ 
& \# Epochs & 10 & 10 & 30 & 20 & 10 & 6 & 15 & 15 \\ \midrule
\multirow{ 2}{*}{\textsc{SVFT}$^{P}$} & Batch Size & 32 & 32 & 16 & 16 & 32 & 16 & 4 & 32 \\
& Learning Rate & 5E-02 & 5E-02 & 5E-02 & 8E-02 & 8E-02 & 5E-02 & 5E-02 & 5E-02  \\ \midrule
\multirow{ 2}{*}{\textsc{SVFT$^{R}_{d=2}$}} & Batch Size & 32 & 32 & 16 & 16 & 32 & 32 & 16 & 32 \\
& Learning Rate & 1E-02 & 1E-02 & 1E-02 & 1E-02 & 3E-02 & 1E-02 & 3E-02 & 1E-02 \\
\bottomrule
\end{tabular}
}
\label{tab:glue_deberta_hparams}
\end{table*}

\paragraph{Natural Language Generation.} See the hyperparameters used in our experiments in ~\autoref{tab:math_hparams}. For LoRA, DoRA, we adapt \(Q, K, V, U, D\) matrices. We apply BOFT on \(Q, V\) matrices since applying on multiple modules is computationally expensive. For VeRA, which enforces a constraint of uniform internal dimensions for shared matrices, we apply on \(G, U\) projection matrices as it yields the highest number of learnable parameters. We apply SVFT on \(Q, K, V, U, D, O, G\) for the Gemma family of models, and \(U, D, O, G\) for LLaMA-3-8B. Note that applying \textsc{SVFT} on these modules does not increase trainable parameters at the same rate as applying LoRA or DoRA on them would. We adopt the code base from \url{https://github.com/meta-math/MetaMath.git} for training scripts and evaluation setups and use the fine-tuning data available at \url{https://huggingface.co/datasets/meta-math/MetaMathQA-40K}.
\begin{table*}[h]
\centering
\caption{Hyperparameter setup used for fine-tuning on MetaMathQA-40K.}
\begin{tabular}{lcccccc}\toprule
\multirow{ 2}{*}{\textbf{Hyperparameter}} & \multicolumn{2}{c}{Gemma-2B} & \multicolumn{2}{c}{Gemma-7B} & \multicolumn{2}{c}{LLaMA-3-8B} \\ \cmidrule(l){2-3} \cmidrule(l){4-5} \cmidrule(l){6-7} 
 & \textsc{SVFT}$^{P}$ & \textsc{SVFT}$^{R}_{d=16}$ & \textsc{SVFT}$^{P}$ & \textsc{SVFT}$^{R}_{d=16}$ & \textsc{SVFT}$^{P}$ & \textsc{SVFT}$^{R}_{d=12}$ \\ \midrule
Optimizer & \multicolumn{6}{c}{AdamW} \\
Warmup Ratio & \multicolumn{6}{c}{0.1} \\
LR Schedule & \multicolumn{6}{c}{Cosine} \\ 
Learning Rate & 5E-02 & 1E-03 & 5E-02 & 1E-03 & 5E-02 & 1E-03 \\ 
Max Seq. Len. & \multicolumn{6}{c}{512}  \\ 
\# Epochs & \multicolumn{6}{c}{2} \\
Batch Size & \multicolumn{6}{c}{64} \\

\bottomrule
\end{tabular}
\label{tab:math_hparams}
\end{table*}

\paragraph{Commonsense Reasoning.} See the hyperparameters used in our experiments in ~\autoref{tab:commonsense_hparams}. We adopt the same set of matrices as that of natural language generation tasks. We use the code base from \url{https://github.com/AGI-Edgerunners/LLM-Adapters}, which also contains the training and evaluation data. 
\begin{table*}[h]
\centering
\caption{Hyperparameter setup used for fine-tuning on commonsense-15K.}
\begin{tabular}{lcccc}\toprule
\multirow{ 2}{*}{\textbf{Hyperparameter}} & \multicolumn{2}{c}{Gemma-2B} & \multicolumn{2}{c}{Gemma-7B}\\ \cmidrule(l){2-3} \cmidrule(l){4-5} 
 & \textsc{SVFT}$^{P}$ & \textsc{SVFT}$^{B}_{d=8}$ & \textsc{SVFT}$^{P}$ & \textsc{SVFT}$^{B}_{d=8}$ \\ \midrule
Optimizer & \multicolumn{4}{c}{AdamW} \\
Warmup Steps & \multicolumn{4}{c}{100} \\
LR Schedule & \multicolumn{4}{c}{Linear} \\
Max Seq. Len. & \multicolumn{4}{c}{512}  \\ 
\# Epochs & \multicolumn{4}{c}{3} \\
Batch Size & \multicolumn{4}{c}{64}  \\
Learning Rate & 5E-02 & 5E-03 & 5E-02 & 1E-03 \\

\bottomrule
\end{tabular}
\label{tab:commonsense_hparams}
\end{table*}

\subsection{Settings for Vision Tasks}
\label{app:vision}
\begin{table*}[h]
\centering
\caption{Hyperparameter setup used for fine-tuning on all vision tasks.}
\begin{tabular}{lcc} \toprule
\textbf{Hyperparameter} & \multicolumn{1}{c}{ViT-B} & \multicolumn{1}{c}{ViT-L} \\
\midrule
Optimizer & \multicolumn{2}{c}{AdamW} \\
Warmup Ratio & \multicolumn{2}{c}{0.1} \\
Weight Decay & \multicolumn{2}{c}{0.01} \\
LR Schedule & \multicolumn{2}{c}{Linear} \\ 
\# Epochs & \multicolumn{2}{c}{10} \\
Batch Size & \multicolumn{2}{c}{64} \\
$\textsc{SVFT}^P$ Learning Rate (Head) & \multicolumn{2}{c}{4E-03} \\ 
$\textsc{SVFT}^P$ Learning Rate & \multicolumn{2}{c}{5E-02} \\
\( \textsc{SVFT}^B_{d=2} \) Learning Rate (Head) & \multicolumn{2}{c}{4E-03} \\ 
\( \textsc{SVFT}^B_{d=2} \) Learning Rate  & \multicolumn{2}{c}{5E-02} \\
\( \textsc{SVFT}^B_{d=8} \) Learning Rate (Head) & \multicolumn{2}{c}{4E-03} \\ 
\( \textsc{SVFT}^B_{d=8} \) Learning Rate & \multicolumn{2}{c}{5E-03} \\
\bottomrule
\end{tabular}
\label{tab:vision_hparams}
\end{table*}

For each dataset in the vision tasks, we train on 10 samples per class, using 2 examples per class for validation, and test on the full test set.
Similar to previous literature, we always train the classifier head for these methods since the number of classes is large. 
The parameter counts do not include the number of parameters in the classification head.
The hyperparameters are mentioned in ~\autoref{tab:vision_hparams}.
We tune the learning rates for \textsc{SVFT} and BOFT select learning rates for other methods from \cite{vera}, run training for 10 epochs, and report test accuracy for the best validation model.
For all methods, since classification head has to be fully trained, we report the parameter count other than the classification head.


\end{document}